%% file: main.tex
\documentclass[11pt]{article}

\usepackage[preprint]{acl}

\usepackage{times}
\usepackage{latexsym}

\usepackage[T1]{fontenc}

\usepackage[utf8]{inputenc}

\usepackage{microtype}

\usepackage{inconsolata}

\usepackage{graphicx}
\usepackage{amsmath}
\usepackage{amssymb}
\usepackage{booktabs}
\usepackage{multirow}
\usepackage{xcolor}
\usepackage{subcaption}
\newcommand{\jsd}{\mathrm{JSD}}
\newcommand{\kld}{D_{\mathrm{KL}}}
\title{LLMs Capture Emotion Labels, Not Emotion Uncertainty: Distributional Analysis and Calibration of Human--LLM Judgment Gaps}

\author{
  \textbf{Keito Inoshita\textsuperscript{1}},
  \textbf{Xiaokang Zhou\textsuperscript{2,3}},
  \textbf{Akira Kawai\textsuperscript{4,5}},
  \textbf{Katsutoshi Yada\textsuperscript{2}}
  \\
  \\
  \textsuperscript{1}Faculty of Business and Commerce, Kansai University, Osaka, Japan \\
  \textsuperscript{2}Faculty of Business Data Science, Kansai University, Osaka, Japan \\
  \textsuperscript{3}RIKEN Center for Advanced Intelligence Project, Tokyo, Japan \\
  \textsuperscript{4}Faculty of Data Science, Shiga University, Hikone, Japan \\
  \textsuperscript{5}Japan Safety Society Research Center \\
  \small{
    \texttt{k845578@kansai-u.ac.jp},
    \texttt{zhou@kansai-u.ac.jp},
    \texttt{akira-kawai@biwako.shiga-u.ac.jp},
    \texttt{yada@kansai-u.ac.jp}
  }
}

\begin{document}
\maketitle
\begin{abstract}
Human annotators frequently disagree on emotion labels, yet most evaluations of Large Language Model (LLM) emotion annotation collapse these judgments into a single gold standard, discarding the distributional information that disagreement encodes.
We ask whether LLMs capture the structure of this disagreement, not just majority labels, by comparing emotion judgment distributions between human annotators and four zero-shot LLMs, plus a fine-tuned RoBERTa baseline, across two complementary benchmarks: GoEmotions and EmoBank, totaling 640{,}000 LLM responses.
Zero-shot models diverge substantially from human distributions, and in-domain fine-tuning, not model scale, is required to close the gap.
We formalize a lexical-grounding gradient through a quantitative transparency score that predicts per-category human--LLM agreement: LLMs reliably capture emotions with explicit lexical markers but systematically fail on pragmatically complex emotions requiring contextual inference, a pattern that replicates across both categorical and continuous emotion frameworks.
We further propose three lightweight post-hoc calibration methods that reduce the distributional gap by up to 14\%, and provide actionable guidelines for when LLM emotion annotations can, and cannot, substitute for human labeling.
\end{abstract}

\input{documents/introduction}
\input{documents/related_work}
\input{documents/methodology}
\input{documents/experiments}
\input{documents/discussion}
\input{documents/conclusion}

\section*{Limitations}
\label{sec:limitations}
\input{documents/limitations}

\section*{Ethics Statement}
\label{sec:ethics}
\input{documents/ethics}

\bibliography{references}

\appendix
\input{documents/appendix}

\end{document}

%% file: documents/introduction.tex

\section{Introduction}
\label{sec:introduction}

When multiple humans label the emotion of a text, they often disagree.
A sarcastic Reddit comment may be judged as amusement by one annotator, annoyance by another, and neutral by a third.
Far from being noise, such disagreement reflects genuine differences in emotional perception, what \citet{1} terms Human Label Variation (HLV).

As Large Language Models (LLMs) are increasingly deployed as emotion annotators \citep{2,3}, a natural question arises: do LLMs exhibit the same patterns of uncertainty as human annotators?
Most prior work evaluates LLM emotion annotation against aggregated human labels (majority vote), discarding the distributional information that makes emotion annotation interesting \citep{4,5}.

\citet{6} evaluate how reasoning settings affect LLMs' ability to model human annotation disagreement, measuring variance correlation and distributional alignment across binary classification subtasks.
Our work differs in three key respects: i)~we analyze the full 28-category emotion distribution rather than reducing to binary subtasks, enabling fine-grained per-category structural analysis; ii)~we go beyond evaluation to propose post-hoc calibration methods that actively reduce the human--LLM distributional gap; and iii)~we formalize a quantitative lexical transparency score that predicts which emotion categories are amenable to LLM annotation.

Concretely, we treat emotion judgments from both humans and LLMs as probability distributions and compare their structures across two complementary emotion frameworks, categorical labels and continuous ratings, using multiple zero-shot LLMs alongside a fine-tuned baseline.
This leads to the following research questions:
\begin{description}
    \item[RQ1] How do human and LLM emotion distributions structurally differ?
    \item[RQ2] When human emotion judgment uncertainty is high, does LLM uncertainty also increase?
    \item[RQ3] Do divergence patterns vary across emotion categories, and can the variation be predicted by a quantitative lexical transparency score?
    \item[RQ4] Can post-hoc calibration methods reduce the distributional gap?
\end{description}

Our main contributions are as follows:

\begin{itemize}
    \item[i)] A distributional analysis of zero-shot LLM emotion judgments across four model families reveals substantial structural divergence from human label distributions, and a novel lexical transparency score combining embedding similarity and lexicon coverage is shown to  
  significantly predict which emotion categories are amenable to reliable LLM annotation.
    \item[ii)] An uncertainty correspondence analysis comparing human annotator disagreement entropy with LLM output entropy uncovers a structural ceiling in current zero-shot models that in-domain supervised fine-tuning substantially mitigates.           
    \item[iii)] Three lightweight post-hoc calibration methods  are proposed and systematically compared, yielding actionable guidelines for practitioners on when and how LLM emotion annotations can be deployed as a cost-effective substitute for human labeling. 
\end{itemize}

%% file: documents/related_work.tex

\section{Related Work}
\label{sec:related_work}

\subsection{Human Label Variation and Emotion Annotation}
Annotator disagreement was traditionally treated as noise \citep{7}, but \citet{8} showed that preserving annotator-level information improves model calibration.
\citet{1} formalized HLV, arguing that disagreement reflects genuine interpretive differences rather than error.
\citet{9} demonstrated that apparent disagreement in textual inference reflects genuine ambiguity, and \citet{10} showed that modeling disagreement explicitly improves performance on subjective tasks.

Recent work traces HLV's evolution from noise to signal \citep{5}, applies it to active learning \citep{11}, and proposes methods to select appropriate agreement metrics \citep{12}.
\citet{13} compared human and model uncertainty in moral foundation annotation, finding that text complexity drives both; however, they focused on fine-tuned models rather than LLMs in zero-shot settings.
\citet{14} argued against standard calibration metrics when human annotators disagree, motivating distributional evaluation approaches.
\citet{15} warned against the ecological fallacy in annotation, showing that individual-level variation is not reducible to group demographics.

GoEmotions \citep{16} provides 58K Reddit comments with 28 emotion categories and preserved per-annotator labels, while EmoBank \citep{17} offers continuous Valence-Arousal-Dominance (VAD) annotations.
Both datasets enable studying label variation, yet most prior work uses only aggregated labels \citep{18,19}.

\subsection{LLMs as Annotators and Disagreement Modeling}
Evaluations of LLMs on emotion tasks find that they excel at simple sentiment but struggle with complex affect \citep{2}, with biases toward negative emotions \citep{20} and systematic differences from human perception \citep{3,21}.
\citet{22} found ChatGPT outperforms crowd workers on text annotation, while \citet{23} urged caution about LLM annotations for social science.
\citet{24} proposed Jury Learning to integrate dissenting annotator voices into models, and \citet{25} explored what can be learned from collectively exhaustive label sets.

Most directly related, \citet{6} investigated whether LLMs can predict human disagreement across multiple tasks, finding that LLMs generally struggle with disagreement modeling; their follow-up further examined reasoning strategies.
\citet{26} proposed alternative agreement measurement approaches for emotion annotations.
While prior work evaluates LLMs as replacement annotators, our work is both descriptive and constructive: we treat human and LLM judgments alike as distributions, compare their uncertainty structures, and propose calibration methods to reduce the gap.

%% file: documents/methodology.tex

\section{Methodology}
\label{sec:methodology}

\begin{figure*}[t]
  \centering
  \includegraphics[width=\linewidth]{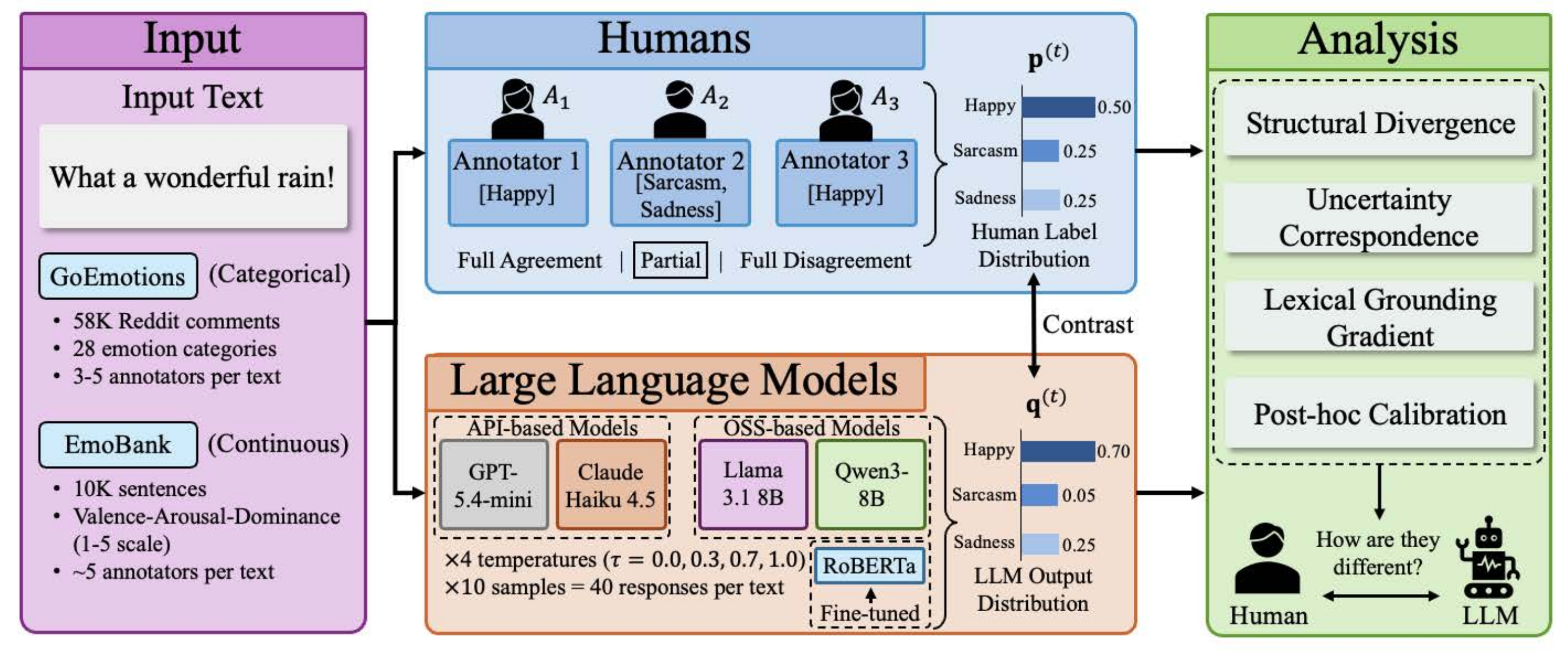}
  \caption{Overview of our experimental framework.}
  \label{fig:framework}
\end{figure*}

We treat both human annotations and LLM outputs as probability distributions over emotion categories, enabling direct structural comparison without privileging either as ground truth. Figure~\ref{fig:framework} illustrates the overall pipeline: input texts are annotated by both human annotators and LLMs, producing two sets of distributions that are then compared through multiple complementary metrics. 

\subsection{Datasets}
\label{sec:dataset}

\paragraph{GoEmotions.}
 We use GoEmotions~\citep{16}, a corpus of 58{,}011 English Reddit comments annotated with 27 emotion categories plus neutral. Each text is labeled by three to five annotators, and the dataset preserves individual annotator judgments. For each text $t$, we construct the human emotion distribution $\mathbf{p}^{(t)} \in \mathbb{R}^{28}$ as the fraction of annotators selecting each category, normalized to sum to 1. 

We classify texts by annotator agreement level into three tiers: full agreement (all annotators select the same emotion; 13.6\%), partial agreement (80.0\%), and full disagreement (no two annotators share any label; 6.3\%). This three-tier classification allows us to systematically analyze how distributional divergence varies across agreement regimes, reflecting the core insight of HLV \citep{1} that annotator disagreement is informative rather than noise. We sample a core set of 2{,}000 texts stratified by agreement level (500/1{,}000/500) to ensure adequate representation of all disagreement regimes. This size was chosen to balance API cost constraints (${\sim}\$400$ total for 640K API calls) with statistical power; bootstrap CIs ($\S$\ref{sec:rq1}) confirm that 2{,}000 texts yield well-separated confidence intervals for our primary metrics.

\paragraph{EmoBank.}
 To validate findings across emotion representation types, we additionally evaluate on EmoBank \citep{17}, a corpus of 10{,}062 English sentences annotated with continuous Valence--Arousal--Dominance (VAD) ratings on a 1--5 scale by 5--7 raters per text. Unlike GoEmotions, where we construct categorical distributions, here we summarize human judgments via the per-text mean and standard deviation of each VAD dimension. Annotator agreement is quantified as the average standard deviation across the three dimensions, and texts are classified into three tiers: high agreement (mean SD $< 0.3$), moderate agreement ($0.3 \leq$ mean SD $< 0.7$), and low agreement (mean SD $\geq 0.7$). We select a core set of 2{,}000 texts stratified by agreement level and valence range, restricted to texts with at least three annotators.                 

\subsection{Models}
\label{sec:models}

We evaluate four zero-shot LLMs and one fine-tuned baseline:

\paragraph{Zero-shot LLMs.}
Two API-based models, GPT-5.4-mini \citep{27} and Claude Haiku 4.5 \citep{28}, and two open-source 8B-class models, Llama~3.1~8B \citep{29} (Meta) and Qwen3-8B \citep{30} (Alibaba), run via vLLM with native chat templates.
This design enables controlled comparison along two dimensions: API vs.\ open-source, and different providers within each category.
Each model receives an identical zero-shot prompt (Appendix~\ref{app:prompt}); for EmoBank, an analogous VAD prediction prompt is used (Appendix~\ref{app:emobank_prompt}).

\paragraph{Fine-tuned baseline.}
To contextualize zero-shot performance, we include RoBERTa-base fine-tuned on GoEmotions \citep{31}, a supervised model trained on the full GoEmotions training set.
Its softmax output provides a probability distribution over all 28 categories without requiring multi-sample aggregation.
This baseline establishes the alignment achievable with in-domain supervised training.

\subsection{LLM Distribution Construction}
\label{sec:distribution}

For each text, we generate responses at four temperatures ($\tau \in \{0.0, 0.3, 0.7, 1.0\}$) with 10 independent samples per temperature, yielding 40 responses per text per model.
Across both datasets and all four models, this produces 640{,}000 total LLM responses.
The LLM emotion distribution $\mathbf{q}^{(t)}$ is constructed as the fraction of samples selecting each emotion, analogous to how human distributions arise from multiple annotators making independent judgments.

We emphasize that temperature-based sampling and human annotator disagreement are conceptually distinct phenomena. Human disagreement arises from genuine differences in emotional perception, cultural background, and interpretive context \citep{15}, whereas temperature sampling modulates the entropy of a model's output distribution.
We do not claim temperature sampling directly simulates human disagreement; rather, we use it as a practical method to approximate the LLM's output distribution over emotion labels, treating the aggregated samples as an empirical estimate of the model's soft predictions.
To address this concern, we report per-temperature results alongside aggregate results.

\subsection{Evaluation Metrics}
\label{sec:metrics}

\paragraph{Jensen--Shannon Divergence.}
Our primary metric is the Jensen--Shannon divergence (JSD) \citep{32}, which is symmetric and, with log base 2, bounded in $[0,1]$:
\begin{equation}
    \jsd(\mathbf{p}, \mathbf{q})
    = \frac{1}{2}\kld(\mathbf{p}\|\mathbf{m})
    + \frac{1}{2}\kld(\mathbf{q}\|\mathbf{m}),
\end{equation}
where $\mathbf{m} = \frac{1}{2}(\mathbf{p}+\mathbf{q})$.
We additionally report the KL divergence and the Wasserstein distance.

\paragraph{Uncertainty quantification.}
We use Shannon entropy, $H(\mathbf{p}) = -\sum_k p_k \log_2 p_k$, to quantify the uncertainty of each distribution. We then compute Spearman's $\rho$ between human and LLM entropies across texts to measure uncertainty correspondence.

\paragraph{Per-category profiling.}
For each emotion category $k$, we compute the mean rate difference, $\Delta_k = \bar{q}_k - \bar{p}_k$, where positive values indicate LLM over-prediction. We also compute, for each category, the Spearman correlation between human and LLM rates across texts.

\paragraph{Lexical transparency score.}
We quantify the lexical grounding of each emotion category using two complementary measures: i)~embedding similarity: cosine similarity between the emotion label embedding and the mean embedding of positively-labeled texts (using sentence-transformers); ii)~lexicon coverage: fraction of positively-labeled texts containing at least one word from the NRC Emotion Lexicon \citep{33} for that category.
Both scores are min-max normalized and averaged to produce a combined transparency score per category.

\paragraph{Statistical testing.}
We use Kruskal--Wallis $H$-tests with Dunn's post-hoc correction for agreement-level comparisons, bootstrap resampling (1{,}000 iterations) for 95\% CIs, and both Cohen's $d$ and Cliff's $\delta$ for pairwise effect sizes.

%% file: documents/experiments.tex

\section{Experiments and Results}
\label{sec:experiments}

We evaluate all models on the GoEmotions and EmoBank core sets described in Section~\ref{sec:methodology}, collecting a total of 640{,}000 LLM responses across both datasets. Below, we address each research question in turn.     

\subsection{RQ1: Structural Differences}
\label{sec:rq1}

Table~\ref{tab:overall_jsd} reports JSD between human and LLM emotion distributions, where each LLM distribution is constructed by aggregating all 40 responses per text across temperatures and samples. All zero-shot models show substantial divergence (JSD $\geq 0.45$), while the fine-tuned RoBERTa baseline achieves roughly half the gap. Among zero-shot models, Qwen3-8B achieves the lowest JSD, followed by GPT-5.4-mini, Llama~3.1~8B, and Claude Haiku~4.5.  

\begin{table}[t]
\centering
\small
\begin{tabular}{lccc}
\toprule
\textbf{Model} & \textbf{JSD}$_\mu$ & \textbf{JSD}$_\sigma$ & \textbf{JSD}$_{\text{med}}$ \\
\midrule
RoBERTa-GoEm (FT)   & 0.299 & 0.233 & 0.236 \\
\midrule
GPT-5.4-mini   & 0.558 & 0.325 & 0.546 \\
Claude Haiku 4.5   & 0.587 & 0.305 & 0.579 \\
Llama 3.1 8B   & 0.584 & 0.278 & 0.566 \\
Qwen3-8B       & 0.453 & 0.279 & 0.454 \\
\bottomrule
\end{tabular}
\caption{Aggregate JSD between human and model emotion distributions.}
\label{tab:overall_jsd}
\end{table}

\begin{figure*}[t]
    \centering
    \includegraphics[width=0.8\linewidth]{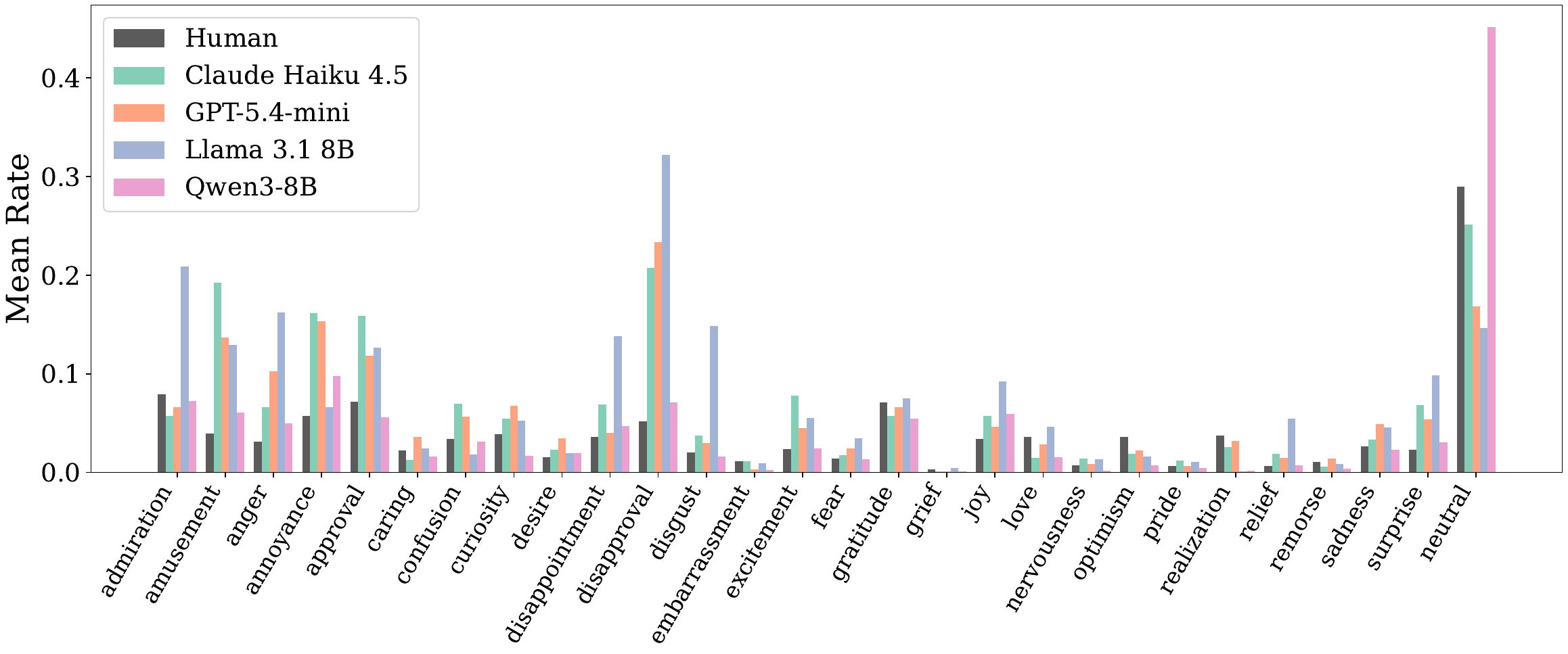}
    \caption{Marginal emotion frequencies: humans vs.\ LLMs.}
    \label{fig:marginal}
\end{figure*}

Figure~\ref{fig:marginal} reveals qualitatively different bias profiles: GPT-5.4-mini, Claude Haiku 4.5, and Llama~3.1~8B over-predict negative emotions (disapproval $+0.16$--$0.27$, annoyance $+0.10$) and under-predict neutral ($-0.12$ to $-0.14$), while Qwen3-8B shows more moderate neutral over-prediction ($+0.16$) with a relatively balanced profile.

\paragraph{Per-temperature results.}
Table~\ref{tab:temp_detail} reports JSD and entropy correlation ($\rho$) at each temperature. JSD decreases monotonically with temperature for all models, confirming that higher sampling diversity improves aggregate distributional alignment. Notably, the magnitude of this improvement differs substantially: open-source models exhibit 2--3$\times$ greater temperature sensitivity than API models.

Entropy correlation reveals a more nuanced pattern. At $\tau = 0$, API and open-source models show a clear gap: GPT-5.4-mini achieves $\rho = 0.216$, whereas the open-source models range from $\rho = 0.114$ to $0.161$. As temperature increases, this gap closes. This convergence is driven primarily by the open-source models, whose uncertainty tracking improves substantially with temperature, while GPT-5.4-mini remains stable.
These results suggest that greedy decoding suppresses the uncertainty signal in open-source models more than in API models, and that temperature-based sampling partially recovers it.

\begin{table}[t]
\centering
\small
\begin{tabular}{llcc}
\toprule
\textbf{Model} & $\tau$ & \textbf{JSD}$_\mu$ & \textbf{Entropy} $\rho$ \\
\midrule
\multirow{4}{*}{GPT-5.4-mini}
 & 0.0 & 0.575 & 0.216 \\
 & 0.3 & 0.573 & 0.213 \\
 & 0.7 & 0.559 & 0.219 \\
 & 1.0 & 0.554 & 0.209 \\
\midrule
\multirow{4}{*}{Claude Haiku 4.5}
 & 0.0 & 0.624 & 0.146 \\
 & 0.3 & 0.609 & 0.179 \\
 & 0.7 & 0.588 & 0.197 \\
 & 1.0 & 0.576 & 0.207 \\
\midrule
\multirow{4}{*}{Llama 3.1 8B}
 & 0.0 & 0.636 & 0.161 \\
 & 0.3 & 0.610 & 0.189 \\
 & 0.7 & 0.592 & 0.222 \\
 & 1.0 & 0.568 & 0.203 \\
\midrule
\multirow{4}{*}{Qwen3-8B}
 & 0.0 & 0.547 & 0.114 \\
 & 0.3 & 0.471 & 0.209 \\
 & 0.7 & 0.457 & 0.211 \\
 & 1.0 & 0.454 & 0.210 \\
\bottomrule
\end{tabular}
\caption{%
  Mean JSD and Spearman entropy correlation ($\rho$) at each temperature $\tau$.
}
\label{tab:temp_detail}
\end{table}

\subsection{RQ2: Uncertainty Correspondence}
\label{sec:rq2}

Table~\ref{tab:entropy_corr} reports Spearman's $\rho$ between per-text human and model entropies.
All zero-shot models show weak but statistically significant correlations, while the fine-tuned baseline achieves substantially higher correspondence.

\begin{table}[t]
\centering
\small
\begin{tabular}{lcc}
\toprule
\textbf{Model} & \textbf{Spearman} $\rho$ & $p$-\textbf{value} \\
\midrule
RoBERTa-GoEm (FT)  & 0.471 & $< 10^{-10}$ \\
\midrule
GPT-5.4-mini    & 0.228 & $< 10^{-10}$ \\
Claude Haiku 4.5   & 0.204 & $< 10^{-10}$ \\
Llama 3.1 8B   & 0.219 & $< 10^{-10}$ \\
Qwen3-8B       & 0.235 & $< 10^{-10}$ \\
\bottomrule
\end{tabular}
\caption{Entropy correlation between human and model distributions.}
\label{tab:entropy_corr}
\end{table}

\begin{figure*}[t]
    \centering
    \includegraphics[width=\linewidth]{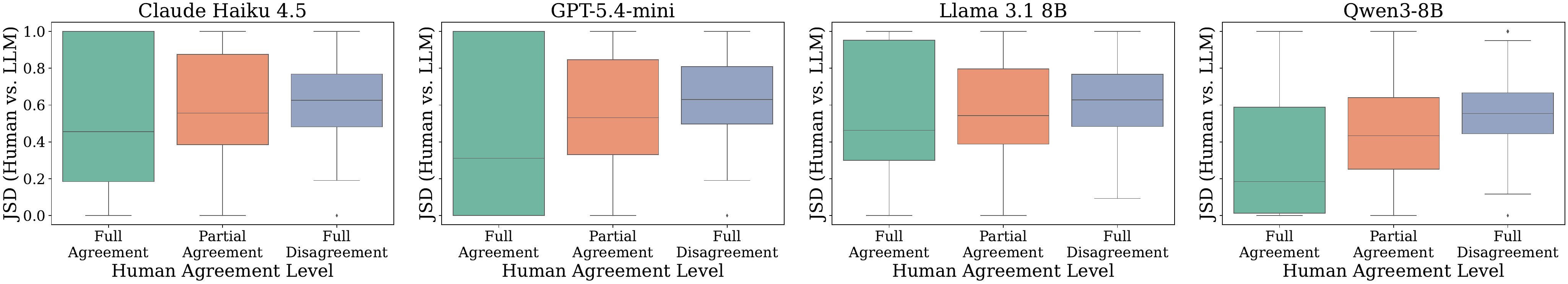}
    \caption{JSD by human agreement level.}
    \label{fig:jsd_agreement}
\end{figure*}

Figure~\ref{fig:jsd_agreement} shows JSD stratified by human agreement level. For all zero-shot models, JSD increases monotonically from full agreement to full disagreement. The fine-tuned RoBERTa baseline (not shown in the figure) exhibits the same monotonic pattern,
but at a substantially lower level, achieving JSD~$0.10$ on full-agreement texts compared to $0.31$--$0.56$ for zero-shot models, indicating that supervised training primarily improves handling of unambiguous cases.

\subsection{RQ3: Emotion-Specific Patterns and Lexical Transparency}
\label{sec:rq3}

\paragraph{Per-category results.}
The per-category analysis (Figure~\ref{fig:cat_corr}; full results in Appendix~\ref{app:category_results}) reveals a consistent gradient across all models: gratitude ($\rho = 0.59$--$0.76$) and love ($\rho = 0.48$--$0.66$) show the highest human--LLM correlations, while approval ($\rho = 0.10$--$0.15$) and realization ($\rho = 0.05$--$0.18$) show the weakest. Per-category sample sizes range from $n=23$ (grief) to $n=451$ (neutral); categories with fewer than 50 positive instances (grief, relief, pride) should be interpreted with caution.

\begin{figure*}[t]
    \centering
    \includegraphics[width=\linewidth]{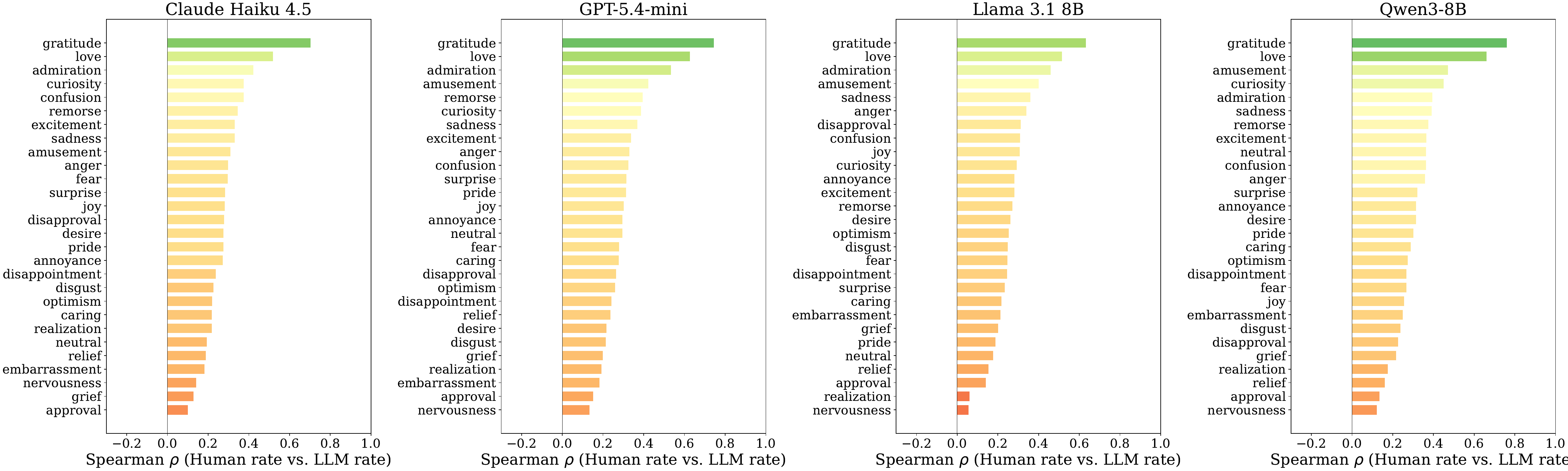}
    \caption{Per-category Spearman $\rho$ (human vs.\ LLM rates).}
    \label{fig:cat_corr}
\end{figure*}

\paragraph{Quantifying the lexical-grounding gradient.}
To move beyond qualitative categorization, we compute a lexical transparency score for each emotion category. Table~\ref{tab:lexical_gradient} shows the Spearman correlation between transparency scores and per-category human--LLM $\rho$. The combined score significantly predicts human--LLM agreement ($r_s = 0.51$, $p = 0.005$, $n = 28$), confirming that the gradient is a quantifiable property rather than a post-hoc narrative.
The top-5 most transparent categories (love, gratitude, remorse, joy, sadness) have mean $\rho = 0.47$, while the bottom-5 (neutral, realization, approval, relief, disapproval) have mean $\rho = 0.19$. Excluding low-frequency categories ($n < 50$: grief, relief) yields a comparable correlation ($r_s = 0.47$, $p = 0.017$, $n = 26$), confirming that the gradient is robust to small-sample categories.

\begin{table}[t]
\centering
\small
\begin{tabular}{lcc}
\toprule
\textbf{Score type} & $r_s$ & $p$-\textbf{value} \\
\midrule
Embedding similarity & 0.424 & 0.025 \\
Lexicon coverage & 0.277 & 0.154 \\
Combined (avg.) & 0.514 & 0.005 \\
\bottomrule
\end{tabular}
\caption{Spearman correlation between lexical transparency score and mean per-category human--LLM $\rho$ ($n=28$ categories).}
\label{tab:lexical_gradient}
\end{table}

\paragraph{API vs.\ open-source.}
OSS models achieve lower aggregate JSD ($0.519$) than API models ($0.573$; Mann--Whitney $p < 10^{-16}$), driven by Qwen3-8B.
However, both categories achieve comparable entropy correlations ($\rho = 0.20$--$0.24$) under standardized prompting.
Bootstrap 95\% CIs confirm robust differences: Qwen3-8B's JSD CI $[0.442, 0.465]$ is clearly separated from the other models (Appendix~\ref{app:bootstrap}).

\subsection{Toward Distributional Calibration}
\label{sec:calibration}

Given the persistent distributional gap, we evaluate three post-hoc calibration methods to improve zero-shot LLM alignment with human distributions, following calibration principles from \citet{34} and distributional considerations from \citet{14}.

\paragraph{Methods.}
Using 5-fold cross-validation on the 2{,}000-text core set:
\begin{enumerate}
    \item Temperature scaling: Learn a scalar $T$ to re-scale logits ($\hat{q}_k \propto \exp(z_k / T)$), optimizing mean JSD on the training fold.
    \item Bias correction: Estimate per-category bias $b_k = \bar{q}_k - \bar{p}_k$ on training data, subtract from predictions, and re-normalize.
    \item Isotonic regression: Fit a non-parametric monotone mapping per category between LLM and human rates \citep{34}.
\end{enumerate}

\paragraph{Results.}
Table~\ref{tab:calibration} summarizes calibration results across all four zero-shot models. Isotonic regression consistently achieves the largest JSD reduction, while also improving entropy correlation. Bias correction is effective for models with strong systematic biases (GPT: $-9.5\%$, Llama: $-11.3\%$) but can increase JSD for Qwen3-8B, whose bias profile is more balanced.

Paired Wilcoxon signed-rank tests confirm that isotonic regression significantly reduces per-text JSD for GPT ($p < 10^{-31}$, $r = 0.26$), Claude ($p < 10^{-38}$, $r = 0.29$), and Llama ($p < 10^{-53}$, $r = 0.35$), with 62--64\% of texts showing improvement; for Qwen3-8B, the effect reverses ($p < 10^{-3}$, $r = 0.08$), consistent with its already balanced profile. Even after calibration, the best zero-shot model remains substantially above the fine-tuned baseline.

\begin{table}[t]
\centering
\small
\setlength{\tabcolsep}{3pt}
\begin{tabular}{lcccc}
\toprule
\textbf{Method} & \textbf{GPT} & \textbf{Claude} & \textbf{Llama} & \textbf{Qwen} \\
\midrule
Uncalibrated       & 0.558 & 0.587 & 0.584 & 0.453 \\
Temp.\ scaling     & 0.546 & 0.562 & 0.572 & 0.449 \\
Bias correction    & 0.505 & 0.559 & 0.518 & 0.487 \\
Isotonic regr.     & 0.491 & 0.518 & 0.500 & 0.471 \\
\midrule
Best $\Delta$JSD   & $-$12.0\% & $-$11.8\% & $-$14.4\% & ---  \\
\bottomrule
\end{tabular}
\caption{Mean JSD after calibration (5-fold CV).}
\label{tab:calibration}
\end{table}

\subsection{Cross-Dataset Validation: EmoBank}
\label{sec:emobank}

We replicate core analyses on EmoBank (2{,}000 texts, continuous VAD, 320{,}000 LLM responses).
Table~\ref{tab:emobank_summary} summarizes the key results.
Qwen3-8B achieves the lowest Mean Absolute Error (MAE) but API models achieve higher Pearson correlations, mirroring the GoEmotions dissociation between aggregate metrics and structural tracking.

\begin{table}[t]
\centering
\small
\begin{tabular}{lcccc}
\toprule
\textbf{Model} & \textbf{MAE} & \textbf{V} $r$ & \textbf{A} $r$ & \textbf{D} $r$ \\
\midrule
GPT-5.4-mini   & 0.512 & 0.669 & 0.342 & 0.336 \\
Claude Haiku 4.5  & 0.548 & 0.658 & 0.342 & 0.280 \\
Llama 3.1 8B   & 0.451 & 0.562 & 0.227 & 0.206 \\
Qwen3-8B       & 0.412 & 0.486 & 0.188 & 0.193 \\
\bottomrule
\end{tabular}
\caption{EmoBank summary: overall MAE and Pearson $r$ by VAD dimension.}
\label{tab:emobank_summary}
\end{table}

Three key patterns replicate: i)~Qwen3-8B achieves best aggregate metrics through compressed predictions (VAD std $= 0.24$--$0.34$ vs.\ human $0.48$--$0.68$) rather than genuine alignment; ii)~API models achieve stronger correlations; iii)~Valence ($r = 0.49$--$0.67$) is predicted far better than Arousal and Dominance ($r = 0.19$--$0.34$), paralleling the lexical-grounding gradient (Table~\ref{tab:cross_dataset}).

\begin{table}[t]
\centering
\small
\setlength{\tabcolsep}{4pt}
\begin{tabular}{lcc}
\toprule
\textbf{Finding} & \textbf{GoEmotions} & \textbf{EmoBank} \\
\midrule
Best aggregate metric & Qwen3-8B & Qwen3-8B \\
Best correlation/rank & GPT-5.4-mini & GPT-5.4-mini \\
Agreement gradient & Monotonic & Monotonic \\
Qwen3-8B strategy & Neutral bias & Mid-point bias \\
Dimension gradient & Lex.\ $>$ Prag. & V $>$ A $\approx$ D \\
\bottomrule
\end{tabular}
\caption{Cross-dataset consistency of key findings.}
\label{tab:cross_dataset}
\end{table}

%% file: documents/discussion.tex

\section{Discussion}
\label{sec:discussion}

\paragraph{A structural gap that calibration alone cannot close.}
Our results reveal that the divergence between human and LLM emotion judgments is not merely a matter of label accuracy but a structural mismatch in how uncertainty is distributed.
All zero-shot LLMs achieve statistically significant entropy correlations with human annotators, yet this correspondence is far too weak to serve as a reliable proxy for human label variation.
The fine-tuned RoBERTa baseline roughly doubles this correspondence while halving aggregate JSD, demonstrating that in-domain supervision, not model scale or prompt engineering, is needed to internalize human uncertainty patterns.

The fact that isotonic regression can reduce JSD by 8--14\% confirms that a monotonic mapping between LLM and human predictions exists: LLMs do preserve ordinal information about emotion intensity.
However, even the best calibrated zero-shot model remains far above the fine-tuned baseline, indicating that post-hoc correction recovers surface-level correspondences but cannot compensate for the absence of learned distributional structure.
The unclosed gap represents aspects of human emotion perception (contextual inference, pragmatic reasoning, cultural common ground) that current zero-shot LLMs do not capture.

\paragraph{What LLMs read vs.\ what they miss: the lexical-grounding gradient.}
The lexical transparency score provides a principled explanation for why LLMs succeed on some emotions and fail on others.
Emotions with explicit lexical markers---gratitude (``thank''), love (``love''), sadness (``sad'')---are predicted well because LLMs can rely on surface-level pattern matching.
Pragmatically complex emotions like approval and realization, which require inference about speaker intent and situational context, consistently show the weakest alignment.

This gradient replicates across both datasets: in EmoBank, Valence (the most lexically transparent VAD dimension) is predicted far better than Arousal and Dominance, which require deeper contextual understanding.
The practical implication is clear: the transparency score can serve as an a priori indicator of whether a given emotion category is safe to annotate with LLMs, enabling practitioners to allocate human review resources where they are most needed.

\paragraph{Model-specific failure modes and the limits of aggregate metrics.}
The four zero-shot models exhibit qualitatively distinct failure modes that aggregate metrics alone cannot reveal.
GPT-5.4-mini, Claude Haiku~4.5, and Llama~3.1~8B share a systematic negative-emotion bias, over-predicting disapproval and under-predicting neutral. Qwen3-8B, by contrast, adopts a conservative strategy of moderate neutral over-prediction with a relatively balanced profile across other categories. This conservatism produces the best aggregate JSD precisely because it avoids large errors, not because it captures human distributions more faithfully.

The same dissociation appears in EmoBank, where Qwen3-8B achieves the lowest MAE through compressed prediction variance rather than genuine alignment with human ratings.
These observations underscore that a single aggregate metric can be misleading: low JSD or MAE does not guarantee that an LLM has learned the structure of human emotion judgment.
Researchers evaluating LLMs as annotators should therefore report per-category profiles and uncertainty correspondence alongside aggregate scores.

\paragraph{Implications for NLP practitioners.}
Based on our findings across two datasets and five models, we offer the following guidelines:
i)~Before deploying LLMs as emotion annotators, compute per-category human--LLM correlations; categories with $\rho < 0.3$ should not be annotated without human validation.
ii)~When a small labeled development set is available, apply isotonic regression calibration; it reduces JSD at negligible computational cost and is effective for all models with systematic biases.
iii)~Always report distributional metrics (entropy correlation, per-category profiles) alongside aggregate accuracy, as aggregate scores can mask fundamental structural misalignment.
iv)~When in-domain labeled data exists, prefer fine-tuned models: even a modest-sized fine-tuned model substantially outperforms much larger zero-shot LLMs on distributional alignment.

%% file: documents/conclusion.tex

\section{Conclusion}
\label{sec:conclusion}

We compared emotion judgment distributions between human annotators, four zero-shot LLMs, and a fine-tuned RoBERTa baseline across GoEmotions and EmoBank, totaling 640{,}000 LLM responses. Zero-shot models show substantial structural divergence with weak uncertainty correspondence, while the fine-tuned baseline roughly halves this gap. Post-hoc isotonic regression reduces zero-shot JSD by up to 14\%, but the calibrated models still fall well short of supervised performance.

Three insights emerge. First, aggregate metrics can mask structural misalignment: the best-performing model by JSD achieves low error through conservative predictions rather than genuine distributional alignment. Second, a lexical transparency score significantly predicts which emotion categories LLMs can reliably annotate, with the same gradient replicating across both datasets. Third, distinct model-specific failure modes persist across datasets, indicating biases rooted in training rather than task design.

Future work should explore larger open-source models, chain-of-thought prompting, disagreement-aware fine-tuning, and cross-lingual validation to further close the gap between LLM and human uncertainty structures.

%% file: documents/limitations.tex

Both datasets are English-only; generalization to other languages and cultural contexts is unverified.
We use zero-shot prompting exclusively for the LLM comparison; few-shot or chain-of-thought prompting may yield different distributional properties.
Our open-source models are limited to the 8B parameter class; larger variants (e.g., 70B+) may show narrower divergence from human distributions.
The sampling-based distribution construction is an approximation of model uncertainty rather than a direct measurement of internal representations \citep{6}.
Our calibration methods are post-hoc and require a labeled development set; their effectiveness may vary across domains and languages.
For low-frequency emotion categories (grief: $n=23$, relief: $n=26$, pride: $n=33$ positive instances in our core set), per-category statistics should be interpreted with appropriate caution due to limited sample sizes.
Finally, we compare at the distributional level but do not analyze individual annotator profiles or demographic factors that may drive disagreement patterns \citep{15}.

%% file: documents/ethics.tex

This study uses publicly available datasets with established ethical approvals.
GoEmotions \citep{16} consists of anonymized Reddit comments released under the Apache 2.0 License, with personally identifiable information removed by the original authors.
EmoBank \citep{17} is released under the CC-BY 4.0 License.
No new human data collection was conducted for this study.

Our work highlights that LLM emotion annotations diverge structurally from human judgments, particularly for pragmatically complex emotions.
We caution against uncritical deployment of LLMs as replacements for human annotators in sensitive applications (e.g., mental health monitoring, content moderation) without careful validation of distributional alignment.
Our calibration methods are intended to improve alignment, not to eliminate the need for human oversight.

API-based model outputs were obtained through standard commercial APIs (OpenAI, Anthropic).
Open-source model inference was conducted on university computing infrastructure.
All model versions and configurations are fully documented to support reproducibility.

%% file: documents/appendix.tex

\onecolumn

\section{GoEmotions Prompt Template}
\label{app:prompt}

All four zero-shot models received identical prompts in a two-message format: a system message defining the task and the set of available emotion labels, followed by a user message containing the target text. For API-based models (GPT-5.4-mini, Claude Haiku 4.5), these were passed as separate system and user roles. For open-source models (Llama~3.1~8B, Qwen3-8B), the same structure was applied via each model's native chat template using vLLM.

\paragraph{System prompt.}
\begin{small}
\begin{verbatim}
You are an emotion annotation assistant. Your task is to identify the
emotions expressed in a given text.

Available emotion labels (select ALL that apply):
admiration, amusement, anger, annoyance, approval, caring, confusion,
curiosity, desire, disappointment, disapproval, disgust, embarrassment,
excitement, fear, gratitude, grief, joy, love, nervousness, optimism,
pride, realization, relief, remorse, sadness, surprise, neutral

Rules:
- Select one or more emotions from the list above.
- If no specific emotion is expressed, select "neutral".
- Return ONLY a JSON array of selected emotion labels, nothing else.
- Example: ["admiration", "joy"]
- Example: ["neutral"]
\end{verbatim}
\end{small}

\paragraph{User prompt.}
\begin{small}
\begin{verbatim}
What emotions are expressed in this text?

Text: "{text}"
\end{verbatim}
\end{small}

\clearpage
\section{EmoBank VAD Prompt Template}
\label{app:emobank_prompt}

For the EmoBank task, we designed a separate prompt to elicit continuous Valence--Arousal--Dominance (VAD) ratings on a 1--5 scale. Following the original EmoBank annotation protocol, the prompt instructs the model to rate from the reader's perspective (i.e., how the text makes the reader feel). The instruction to return a JSON object with three numeric keys ensures machine-parseable output.

\paragraph{System prompt.}
\begin{small}
\begin{verbatim}
You are an emotion annotation assistant. Your task is to rate the
emotional content of a given text on three dimensions:
Valence (unpleasant to pleasant), Arousal (calm to excited), and
Dominance (submissive to dominant).

Rate each dimension on a scale from 1.0 to 5.0, where 3.0 is neutral.

Rules:
- Provide ratings from the READER's perspective
  (how the text makes you feel as a reader).
- Return ONLY a JSON object with three keys: "V", "A", "D"
- Each value must be a number between 1.0 and 5.0
- Example: {"V": 3.2, "A": 2.5, "D": 3.8}
\end{verbatim}
\end{small}

\paragraph{User prompt.}
\begin{small}
\begin{verbatim}
Rate the emotional content of this text on
Valence, Arousal, and Dominance (1.0-5.0):

Text: "{text}"
\end{verbatim}
\end{small}

\clearpage
\section{Full Per-Category Results}
\label{app:category_results}

Table~\ref{tab:full_category} reports the complete per-category statistics for all 28 GoEmotions emotion categories, extending the selective analysis in Section~\ref{sec:rq3}. For each emotion category and each model, we show the rate difference $\Delta$ (LLM prediction rate minus human annotation rate; positive values indicate over-prediction) and the per-text Spearman correlation $\rho$.

Several patterns are consistent across all models. Disapproval is the most over-predicted category in three of four models ($\Delta = +0.182$, $+0.156$, $+0.270$ for GPT, Claude, and Llama respectively), with only Qwen3-8B near-calibrated ($\Delta = +0.020$). The categories with the highest per-text correlations are gratitude ($\rho = 0.632$--$0.760$), love ($\rho = 0.514$--$0.660$), and admiration ($\rho = 0.395$--$0.533$), all emotions with relatively unambiguous lexical markers. The weakest correlations cluster around context-dependent emotions: nervousness ($\rho = 0.056$--$0.128$) and realization ($\rho = 0.061$--$0.175$).

The neutral category shows strikingly divergent behaviour across models: GPT-5.4-mini and Llama~3.1~8B strongly under-predict neutral ($\Delta = -0.122$ and $-0.143$), while Qwen3-8B over-predicts it by the largest margin among all its categories ($\Delta = +0.161$). This contrast directly relates to the model-specific failure modes discussed in Section~\ref{sec:discussion}.

\begin{figure}[h]
    \centering
    \includegraphics[width=\textwidth]{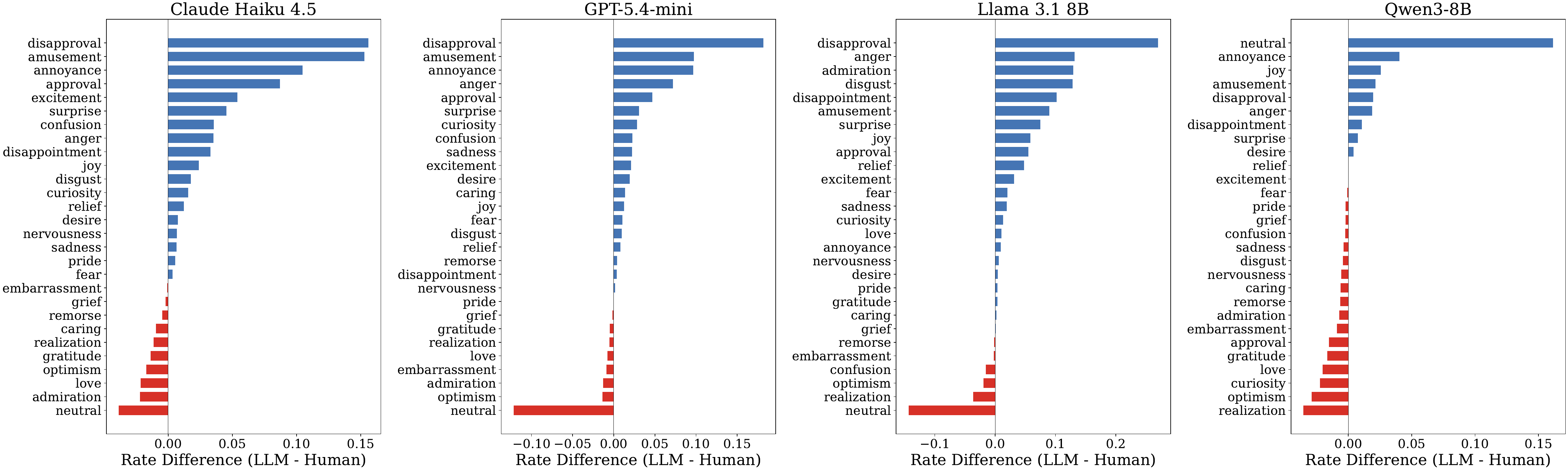}
    \caption{%
      Per-category rate difference ($\Delta$ = LLM $-$ human) for all four models across the 28 GoEmotions categories.
      Bars above zero indicate over-prediction; bars below zero indicate under-prediction.
    }
    \label{fig:rate_diff}
\end{figure}

Figure~\ref{fig:rate_diff} visualises the per-category rate differences for all four models, making the systematic over-prediction of disapproval and the divergent treatment of neutral immediately apparent.

\begin{table*}[h]
\centering
\small
\setlength{\tabcolsep}{3pt}
\begin{tabular}{l|rr|rr|rr|rr}
\toprule
& \multicolumn{2}{c|}{\textbf{GPT-5.4-mini}} & \multicolumn{2}{c|}{\textbf{Claude Haiku 4.5}} & \multicolumn{2}{c|}{\textbf{Llama 3.1 8B}} & \multicolumn{2}{c}{\textbf{Qwen3-8B}} \\
\textbf{Emotion} & $\Delta$ & $\rho$ & $\Delta$ & $\rho$ & $\Delta$ & $\rho$ & $\Delta$ & $\rho$ \\
\midrule
admiration    & $-$0.013 & 0.533 & $-$0.022 & 0.421 & $+$0.130 & 0.459 & $-$0.007 & 0.395 \\
amusement     & $+$0.097 & 0.423 & $+$0.153 & 0.309 & $+$0.090 & 0.400 & $+$0.021 & 0.471 \\
anger         & $+$0.072 & 0.329 & $+$0.044 & 0.300 & $+$0.132 & 0.340 & $+$0.019 & 0.358 \\
annoyance     & $+$0.097 & 0.296 & $+$0.105 & 0.272 & $+$0.010 & 0.282 & $+$0.040 & 0.315 \\
approval      & $+$0.047 & 0.152 & $+$0.087 & 0.101 & $+$0.055 & 0.140 & $-$0.015 & 0.134 \\
caring        & $+$0.029 & 0.252 & $+$0.037 & 0.185 & $+$0.002 & 0.218 & $-$0.006 & 0.289 \\
confusion     & $+$0.010 & 0.378 & $+$0.016 & 0.362 & $-$0.015 & 0.309 & $-$0.003 & 0.363 \\
curiosity     & $+$0.024 & 0.364 & $+$0.042 & 0.282 & $+$0.013 & 0.293 & $-$0.022 & 0.450 \\
desire        & $+$0.010 & 0.283 & $+$0.014 & 0.188 & $+$0.004 & 0.261 & $+$0.004 & 0.315 \\
disappointment& $+$0.031 & 0.202 & $+$0.030 & 0.195 & $+$0.102 & 0.245 & $+$0.011 & 0.267 \\
disapproval   & $+$0.182 & 0.264 & $+$0.156 & 0.278 & $+$0.270 & 0.313 & $+$0.020 & 0.226 \\
disgust       & $+$0.032 & 0.262 & $+$0.025 & 0.293 & $+$0.128 & 0.248 & $-$0.004 & 0.237 \\
embarrassment & $-$0.009 & 0.183 & $+$0.002 & 0.191 & $-$0.002 & 0.212 & $-$0.009 & 0.249 \\
excitement    & $+$0.027 & 0.297 & $+$0.054 & 0.331 & $+$0.031 & 0.280 & $+$0.000 & 0.365 \\
fear          & $+$0.006 & 0.339 & $+$0.010 & 0.341 & $+$0.020 & 0.246 & $-$0.001 & 0.267 \\
gratitude     & $-$0.006 & 0.677 & $-$0.014 & 0.703 & $+$0.004 & 0.632 & $-$0.017 & 0.760 \\
grief         & $+$0.004 & 0.100 & $+$0.003 & 0.101 & $+$0.001 & 0.201 & $-$0.002 & 0.216 \\
joy           & $+$0.012 & 0.425 & $+$0.016 & 0.379 & $+$0.058 & 0.307 & $+$0.026 & 0.256 \\
love          & $-$0.008 & 0.627 & $-$0.021 & 0.518 & $+$0.011 & 0.514 & $-$0.020 & 0.660 \\
nervousness   & $+$0.002 & 0.128 & $+$0.002 & 0.101 & $+$0.006 & 0.056 & $-$0.006 & 0.121 \\
optimism      & $-$0.014 & 0.259 & $-$0.017 & 0.220 & $-$0.019 & 0.253 & $-$0.029 & 0.274 \\
pride         & $+$0.004 & 0.145 & $+$0.004 & 0.146 & $+$0.004 & 0.188 & $-$0.002 & 0.301 \\
realization   & $+$0.010 & 0.124 & $+$0.014 & 0.142 & $-$0.036 & 0.061 & $-$0.035 & 0.175 \\
relief        & $+$0.005 & 0.133 & $+$0.004 & 0.137 & $+$0.048 & 0.154 & $+$0.000 & 0.161 \\
remorse       & $-$0.001 & 0.408 & $+$0.001 & 0.397 & $-$0.002 & 0.271 & $-$0.006 & 0.374 \\
sadness       & $+$0.020 & 0.354 & $+$0.019 & 0.349 & $+$0.019 & 0.360 & $-$0.004 & 0.392 \\
surprise      & $+$0.022 & 0.302 & $+$0.029 & 0.269 & $+$0.075 & 0.234 & $+$0.008 & 0.322 \\
neutral       & $-$0.122 & 0.295 & $-$0.038 & 0.193 & $-$0.143 & 0.177 & $+$0.161 & 0.363 \\
\midrule
\textbf{Mean $|\Delta|$} & \multicolumn{2}{c|}{0.033} & \multicolumn{2}{c|}{0.035} & \multicolumn{2}{c|}{0.051} & \multicolumn{2}{c}{0.018} \\
\textbf{Mean $\rho$}     & \multicolumn{2}{c|}{0.298} & \multicolumn{2}{c|}{0.272} & \multicolumn{2}{c|}{0.273} & \multicolumn{2}{c}{0.324} \\
\bottomrule
\end{tabular}
\caption{%
  Complete per-category results for all 28 GoEmotions emotion categories, aggregated across all temperature settings.
  $\Delta$ = rate difference (LLM $-$ human; positive = over-prediction).
  $\rho$ = per-text Spearman correlation between LLM and human probabilities.
}
\label{tab:full_category}
\end{table*}

\clearpage
\section{Bootstrap Confidence Intervals}
\label{app:bootstrap}

To quantify the statistical reliability of the aggregate metrics reported in Section~\ref{sec:rq1}, we computed 95\% bootstrap confidence intervals using 1{,}000 resampling iterations. Table~\ref{tab:bootstrap} reports CIs for JSD, KL divergence, and entropy correlation. Qwen3-8B's JSD CI $[0.442, 0.465]$ does not overlap with the CIs of any other model (lower bounds $\geq 0.544$), confirming that its lower JSD is not a sampling artefact. The KLD metric reveals a different pattern: both OSS models show substantially lower KLD than the API models, because KLD is sensitive to probability mass near zero and the OSS models concentrate mass on fewer categories.

\begin{table}[h]
\centering
\small
\begin{tabular}{llcc}
\toprule
\textbf{Model} & \textbf{Metric} & \textbf{Mean} & \textbf{95\% CI} \\
\midrule
\multirow{3}{*}{GPT-5.4-mini}
 & JSD & 0.558 & [0.544, 0.571] \\
 & KLD & 19.946 & [19.355, 20.514] \\
 & Entropy $\rho$ & 0.228 & [0.191, 0.271] \\
\midrule
\multirow{3}{*}{Claude Haiku 4.5}
 & JSD & 0.587 & [0.574, 0.600] \\
 & KLD & 19.048 & [18.452, 19.612] \\
 & Entropy $\rho$ & 0.204 & [0.165, 0.248] \\
\midrule
\multirow{3}{*}{Llama 3.1 8B}
 & JSD & 0.584 & [0.571, 0.596] \\
 & KLD & 14.174 & [13.671, 14.672] \\
 & Entropy $\rho$ & 0.219 & [0.178, 0.261] \\
\midrule
\multirow{3}{*}{Qwen3-8B}
 & JSD & 0.453 & [0.442, 0.465] \\
 & KLD & 14.585 & [14.023, 15.159] \\
 & Entropy $\rho$ & 0.235 & [0.196, 0.277] \\
\bottomrule
\end{tabular}
\caption{%
  Bootstrap 95\% confidence intervals (1{,}000 iterations) for key aggregate metrics.
}
\label{tab:bootstrap}
\end{table}

\begin{figure}[h]
    \centering
    \includegraphics[width=\textwidth]{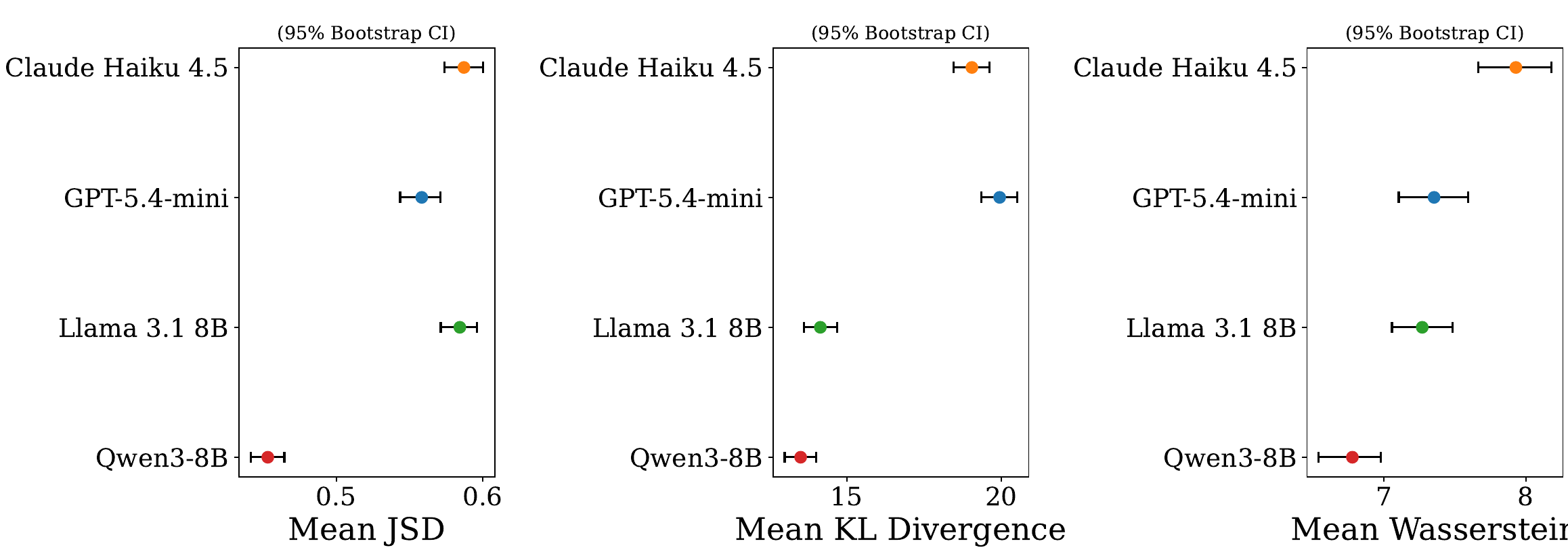}
    \caption{%
      Forest plot of 95\% bootstrap confidence intervals for JSD, KLD, and entropy correlation across all four models.
    }
    \label{fig:bootstrap_forest}
\end{figure}

Figure~\ref{fig:bootstrap_forest} provides a visual comparison of these CIs, making the clear separation of Qwen3-8B from the other three models immediately apparent.

\clearpage
\section{EmoBank Detailed Results}
\label{app:emobank_details}

Table~\ref{tab:emobank_mae} reports the full EmoBank results with 95\% bootstrap CIs and both Pearson and Spearman correlations for each model--dimension combination. All models achieve their best correlations on Valence ($r = 0.486$--$0.669$), consistent with Valence being the most linguistically accessible affective dimension. Arousal proves most challenging (Pearson $r$ as low as $0.188$ for Qwen3-8B), reflecting the difficulty of inferring physiological activation from text alone.

A key observation is that low MAE does not imply high correlation. Qwen3-8B achieves the lowest MAE on Arousal ($0.396$) and Dominance ($0.342$) but the lowest Pearson correlations on both dimensions, because its predictions are compressed towards the mean (prediction std $= 0.24$--$0.34$ vs.\ human std $= 0.48$--$0.68$). This compression reduces absolute errors without capturing genuine within-sample variability.

\begin{table}[h]
\centering
\small
\begin{tabular}{llcccc}
\toprule
\textbf{Model} & \textbf{Dim} & \textbf{MAE} & \textbf{95\% CI} & $r$ & $\rho$ \\
\midrule
\multirow{3}{*}{GPT-5.4-mini}
 & V & 0.514 & [0.494, 0.535] & 0.669 & 0.686 \\
 & A & 0.616 & [0.598, 0.634] & 0.342 & 0.372 \\
 & D & 0.405 & [0.387, 0.423] & 0.336 & 0.381 \\
\midrule
\multirow{3}{*}{Claude Haiku 4.5}
 & V & 0.488 & [0.468, 0.507] & 0.658 & 0.678 \\
 & A & 0.643 & [0.624, 0.663] & 0.342 & 0.374 \\
 & D & 0.511 & [0.493, 0.529] & 0.280 & 0.319 \\
\midrule
\multirow{3}{*}{Llama 3.1 8B}
 & V & 0.444 & [0.427, 0.460] & 0.562 & 0.616 \\
 & A & 0.514 & [0.498, 0.531] & 0.227 & 0.226 \\
 & D & 0.395 & [0.379, 0.410] & 0.206 & 0.246 \\
\midrule
\multirow{3}{*}{Qwen3-8B}
 & V & 0.497 & [0.480, 0.514] & 0.486 & 0.591 \\
 & A & 0.396 & [0.382, 0.411] & 0.188 & 0.258 \\
 & D & 0.342 & [0.328, 0.357] & 0.193 & 0.235 \\
\bottomrule
\end{tabular}
\caption{%
  EmoBank VAD results with 95\% bootstrap CIs, Pearson $r$, and Spearman $\rho$. V = Valence, A = Arousal, D = Dominance.
}
\label{tab:emobank_mae}
\end{table}

\begin{figure}[h]
    \centering
    \includegraphics[width=0.9\textwidth]{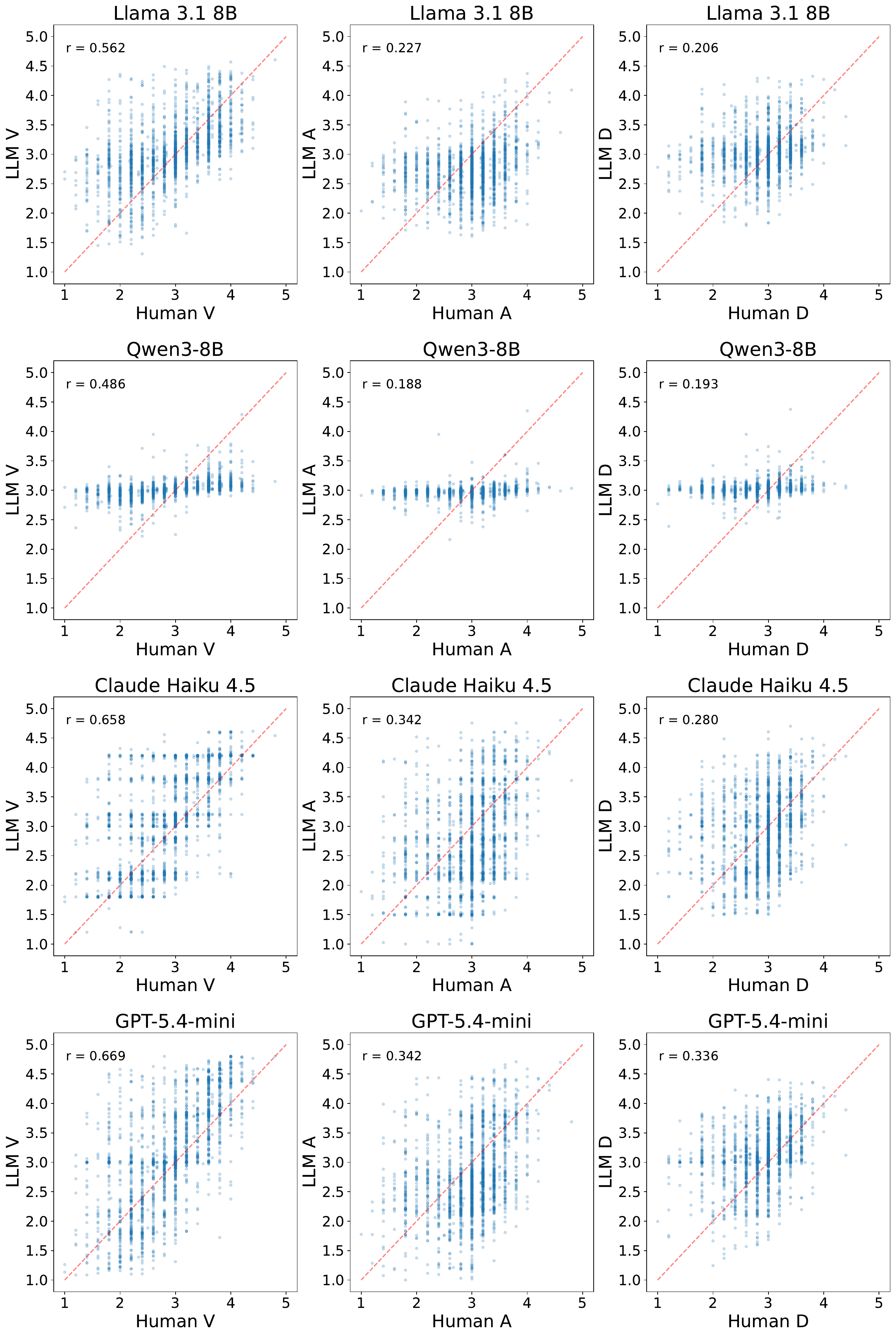}
    \caption{%
      Scatter plots of human vs.\ LLM VAD predictions for each model (rows) and dimension (columns).
      Pearson $r$ is annotated in each panel. The diagonal dashed line indicates perfect agreement.
    }
    \label{fig:emobank_scatter}
\end{figure}

Figure~\ref{fig:emobank_scatter} shows scatter plots of human vs.\ LLM VAD predictions for each model and dimension. The diagonal reference line ($y = x$) highlights the degree of alignment; note how Qwen3-8B's predictions cluster tightly around the mean, producing a visibly compressed range compared to the other models.

\clearpage
\section{Agreement-Level JSD Breakdown}
\label{app:agreement}

Table~\ref{tab:agreement_detail} reports JSD stratified by human inter-annotator agreement level. All models show monotonically increasing JSD from the full agreement tier to the full disagreement tier. Qwen3-8B achieves the lowest JSD in the full agreement tier ($0.314$), indicating strong calibration on unambiguous texts, but this advantage diminishes in the disagreement tier ($0.567$ vs.\ $0.630$--$0.654$ for other models).

\begin{table}[h]
\centering
\small
\begin{tabular}{llccc}
\toprule
\textbf{Model} & \textbf{Agreement} & \textbf{JSD}$_\mu$ & \textbf{JSD}$_\sigma$ & $n$ \\
\midrule
\multirow{3}{*}{GPT-5.4-mini}
 & Full agr. & 0.448 & 0.343 & 500 \\
 & Partial   & 0.566 & 0.312 & 1000 \\
 & Full dis. & 0.654 & 0.286 & 500 \\
\midrule
\multirow{3}{*}{Claude Haiku 4.5}
 & Full agr. & 0.537 & 0.318 & 500 \\
 & Partial   & 0.584 & 0.299 & 1000 \\
 & Full dis. & 0.644 & 0.289 & 500 \\
\midrule
\multirow{3}{*}{Llama 3.1 8B}
 & Full agr. & 0.561 & 0.296 & 500 \\
 & Partial   & 0.573 & 0.268 & 1000 \\
 & Full dis. & 0.630 & 0.261 & 500 \\
\midrule
\multirow{3}{*}{Qwen3-8B}
 & Full agr. & 0.314 & 0.291 & 500 \\
 & Partial   & 0.466 & 0.268 & 1000 \\
 & Full dis. & 0.567 & 0.247 & 500 \\
\bottomrule
\end{tabular}
\caption{%
  JSD by human agreement level.
}
\label{tab:agreement_detail}
\end{table}

\clearpage
\section{Pairwise Effect Sizes}
\label{app:effect_sizes}

Table~\ref{tab:effect_sizes} reports Cohen's $d$, Cliff's $\delta$, and Mann--Whitney $p$-values for all pairwise model comparisons on GoEmotions JSD. The largest effects are between Qwen3-8B and the other models ($d = 0.347$--$0.471$), confirming that Qwen3-8B occupies a distinct performance tier. In contrast, the differences among GPT, Claude, and Llama are small ($d < 0.10$) and two of three comparisons are non-significant, indicating that these three models form a statistically indistinct cluster.

\begin{table}[h]
\centering
\small
\begin{tabular}{lccc}
\toprule
\textbf{Comparison} & \textbf{Cohen's} $d$ & \textbf{Cliff's} $\delta$ & \textbf{MW} $p$ \\
\midrule
Claude vs.\ GPT    & 0.092  & 0.046  & $1.1 \times 10^{-2}$ \\
Claude vs.\ Llama  & 0.010  & 0.021  & $2.5 \times 10^{-1}$ \\
Claude vs.\ Qwen   & 0.458  & 0.249  & $1.7 \times 10^{-42}$ \\
GPT vs.\ Llama     & $-$0.086 & $-$0.029 & $1.2 \times 10^{-1}$ \\
GPT vs.\ Qwen      & 0.347  & 0.187  & $9.7 \times 10^{-25}$ \\
Llama vs.\ Qwen    & 0.471  & 0.252  & $1.9 \times 10^{-43}$ \\
\midrule
API vs.\ OSS       & 0.179  & 0.107  & $9.7 \times 10^{-17}$ \\
\bottomrule
\end{tabular}
\caption{%
  Pairwise effect sizes for GoEmotions JSD. Positive $d$ indicates the first model has higher JSD.
}
\label{tab:effect_sizes}
\end{table}